\documentclass{sigchi-ext}
\usepackage[T1]{fontenc}
\usepackage{textcomp}
\usepackage[scaled=.92]{helvet} % for proper fonts
\usepackage{graphicx} % for EPS use the graphics package instead
\usepackage{balance}  % for useful for balancing the last columns
\usepackage{booktabs} % for pretty table rules
\usepackage{ccicons}  % for Creative Commons citation icons
\usepackage{ragged2e} % for tighter hyphenation

\usepackage[textwidth=1in,textsize=footnotesize]{todonotes}
\usepackage{hyperref}

\usepackage{marginnote} 
\def\plaintitle{LEx: A Framework for Operationalising  Layers of Machine Learning Explanations}

\def\emptyauthor{}
\def\plainkeywords{explainability; XAI; sensitivity; counterfactuals; social considerations}

\title{\textit{LEx}: A Framework for Operationalising  Layers of AI Explanations}

\numberofauthors{4}
\author{%
  \alignauthor{%
   \marginnote{All authors contributed equally to this research.}
    \textbf{Ronal Singh}\\
    \textbf{Marc Cheong}\\
    \textbf{Tim Miller}\\
    \affaddr{School of Computing and Information Systems}\\
    \affaddr{The University of Melbourne}\\
    \email{singhrr@unimelb.edu.au, marc.cheong@unimelb.edu.au, tmiller@unimelb.edu.au} \\
    }\alignauthor{%
    \textbf{Upol Ehsan}\\
    \textbf{Mark O. Riedl}\\
    \affaddr{Georgia Institute of Technology}\\
    \affaddr{Atlanta, Georgia}\\
    \email{ehsanu@gatech.edu, riedl@cc.gatech.edu} } }

\definecolor{linkColor}{RGB}{6,125,233}
\hypersetup{%
  pdftitle={\plaintitle},
  pdfauthor={\emptyauthor},
  pdfkeywords={\plainkeywords},
  bookmarksnumbered,
  pdfstartview={FitH},
  colorlinks,
  citecolor=blue,
  filecolor=black,
  linkcolor=black,
  urlcolor=linkColor,
  breaklinks=true
}

\begin{document}

%% For the camera ready, use the commands provided by the ACM in the Permission Release Form.
\CopyrightYear{2021}
\setcopyright{rightsretained}
\conferenceinfo{ACM CHI Workshop on Operationalizing Human-Centered Perspectives in Explainable AI}{May  8--9, 2021, Virtual}
\isbn{}
\doi{}
%% Then override the default copyright message with the \acmcopyright command.
\copyrightinfo{\acmcopyright}

\maketitle

% Uncomment to disable hyphenation (not recommended)
% https://twitter.com/anjirokhan/status/546046683331973120
\RaggedRight{}

%
% The abstract is a short summary of the work to be presented in the
% article.
\begin{abstract}

Several social factors impact how people respond to AI explanations used to justify AI decisions affecting them personally. In this position paper, we define a framework called the \emph{layers of explanation} (LEx), a lens through which we can assess the appropriateness of different types of explanations. The framework uses the notions of \emph{sensitivity} (emotional responsiveness) of features and the level of \emph{stakes} (decision's consequence) in a domain to determine whether different types of explanations are \emph{appropriate} in a given context. We demonstrate how to use the framework to assess the appropriateness of different types of explanations in different domains.

\end{abstract}

\section{Introduction}

The discourse on explainability in algorithmic decision-making has been gaining traction in the past few years. Research in this area proposes various types of explanation methods, such as feature-based explanations, natural language rationales~\cite{ehsan2019automated}, counterfactual \& contrastive explanations \cite{miller2018contrastive,byrne2019counterfactuals,russell2019efficient}, and directive explanations \cite{singh2021directive}. Despite the recent progress in Explainable AI (XAI), there is an emerging body of work~\cite{alqaraawi2020evaluating,poursabzi2018manipulating,zhang2020effect,ehsan2020human, liao2020questioning, ehsan2020human} that show evidence that explanation systems can fail when they do not consider human factors. If the field of XAI is to address the infamous reputation of ``inmates running the asylum''~\cite{miller2019explanation}, where XAI researchers often develop explanations based on their own intuition rather than the situated needs of their intended audience, we need to go beyond the bounds of the algorithm and incorporate human factors in XAI design~\cite{ehsan2019on, ehsan2021expanding}. While ``opening'' the proverbial black-box of AI matters, \textit{who} opens it and \textit{how} the explanations are given matters just as much, if not more than just the \textit{what}.

Recent studies \cite{Binns2018,singh2021directive} show that certain types of explanations are considered to be \emph{inappropriate} to explain certain aspects of decisions. For example, Singh et al \cite{singh2021directive} show that for explaining credit-scoring decisions, giving a \emph{directive} explanation that an applicant should find a higher-paying job to increase their income (a directive explanation) is `condescending' and `impolite', compared to simply noting that the loan would have been granted \emph{if} the applicant had a higher income (counterfactual explanation). On the other hand, directive explanations were preferred for many other aspects of credit scoring, such as how to optimise the number of credit cards.

Inspired by Nissenbaum's notion of \emph{contextual integrity} \cite{nissenbaum2004privacy}, which argues from a view that privacy protection should be tied to norms within specific contexts, we propose that explanations in AI should be contextualised as well. We propose a framework called the \emph{layers of explanation} (LEx) framework, which operationalizes the \emph{who, why} and more importantly the \emph{how} of explanations from an AI system. In particular, we can use it as an analytic lens to scope the appropriateness of different types of explanations. Inspired by the results of \cite{Binns2018,singh2021directive}, the framework uses the notions of \emph{sensitivity} of features and the level of \emph{stakes} in a domain to determine whether types of explanations are \emph{appropriate} in a given context. We provide examples of how we can apply this framework to assess explanation appropriateness.

\section{Framework: Layers of Explanation (LEx)}

This section defines the rationale behind the LEx framework. The framework proposes three main questions:
\begin{enumerate}
  \item \emph{'Who?'}: Who are the human agents that receive explanations in the domain?

  \item \emph{`Why?'}: Why is the explanation needed and what are the explainability goals?

  \item  \emph{`How'}: How should explanations be given to specific target segments given their explainability goals?

\end{enumerate}

These questions are framed through the lens of how high the stakes are for a given person in a given context, and how sensitive are the features being referred to.

\subsection{Stakes and Sensitivity}

The two concepts used to determine the appropriateness of the explanations are the stakes and sensitivity. The \emph{stakes} are the consequences (positive or negative, that a person receives for obtaining a particular decision, and its impact on their human agency. We divide the stakes into \emph{low}, \emph{medium}, and \emph{high}. A low stakes domain for a person could be creating a personal account on a social media website, while a high stakes domain is applying for a business loan that can make or break one's financial future.

\emph{Sensitivity} is the emotional responsiveness or susceptibility of a person to a particular explanation (or feature of an explanation). We divide sensitivity into \emph{low} and \emph{high}. An example of a low sensitivity feature is informing someone that they cannot purchase a ticket because an event is sold out. An example of a high sensitivity feature is referring to someone's ethnic background, particularly for people in a minority group. 

Note that sensitivity is not the same as a \textit{protected feature}\footnote{See e.g. definitions in the Australian legal context: \url{https://www.fairwork.gov.au/employee-entitlements/protections-at-work/protection-from-discrimination-at-work}}, which is an feature that is (often legally) prevented from being used as part of a decision. A sensitive feature may be legal, but inappropriate. Crenshaw's Intersectionality framework~\cite{Crenshaw1989intersection} may also be relevant from the viewpoint that features interact or overlap with each other. The intersecting combinations of features may directly shape one's circumstance and plays a key role in compounding the sensitivity of sets of feature(s).

Consider the example in the previous section: a sensitive feature may be high for one group, but low for another; ethnicity is a sensitive feature, but may be less sensitive in some domains/contexts for someone with a dominant culture.  Hence, designers of XAI systems, especially of the dominant culture and privileged epistemic positions~\cite{Fricker2007} in society, need to suspend all preconceived knowledge of their own socio-historical and personal circumstances. Table \ref{tab:stakesensitivity} summarises the six potential combinations of sensitivity and stake, along with illustrative examples.

\begin{table}[h]
  \caption{Stake versus feature sensitivity: examples}
  \label{tab:stakesensitivity}
  \begin{tabular}{l p{3cm} p{3cm}}
    \toprule
             & \multicolumn{2}{c}{Feature Sensitivity}\\
             \cmidrule{2-3}
    Stake  & \multicolumn{1}{c}{Lower} & \multicolumn{1}{c}{Higher}\\
    \midrule
    Low & Social media account registration: A user cannot create a new account due to anti-spam heuristics; the user is asked to retry. & Recommender systems: New products are recommended based on a user's purchase history/demographics.\\\hline
    Medium & Exam grading: A student's writing score in an exam is auto-graded based on historical patterns.  & Automated Recruitment: Job outcome for an applicant is due to a protected feature.\\\hline
    High & Loan: A small trader's loan application has been denied due to certain business rule. & Bail: An accused person is denied bail (e.g. due to statistical correlation to a protected feature).\\
  \bottomrule
\end{tabular}
\end{table}

\subsection{Defining the layers}

We propose the \emph{layers of explanation} (see Table 2), inspired by various types of explanations~\cite{guidotti2019survey,adadi2018peeking,lipton2016mythos,molnar2019interpretable}, including Singh et al.'s~\cite{singh2021directive} notion of directive explanation. We also provide examples showing how to develop the layers and assess the appropriateness given prior knowledge of how people may respond to these explanations. 

\marginnote{
\begin{description}
    \item 
        \textbf{Table 2: Defining the layers}
        \label{Table 2}
    \item 
        \textbf{[Baseline] No explanation (decision only)}: The automated systems communicates (only) the decision.
    \item 
        \textbf{[Layer 1] Feature-based explanation}: The explanation states the features relevant to a decision. %There is an implicit assumption that changing these features will change the decision or outcome.
    \item 
        \textbf{[Layer 2] Contrastive explanation}: The explanation provides not only the features and values but also states how the values of the features need to be different for a different outcome or decision.
    \item
        \textbf{[Layer 3] Directive explanation}: The explanation lists all of the information from previous layers and suggests \emph{actions} or \emph{interventions} that could bring about the desired outcome or decision.
        
\end{description}
}[-12.6cm]

Note the ordinal relationship between layers: if one layer is inappropriate or sensitive, any higher layer will also be; e.g. directive explanations refer to counterfactuals, so if a counterfactual explanation is considered insensitive, then so too will the directive explanation.

\hspace{2mm}

\subsection{Process}
In this section, we discuss the process of the LEx framework to judge the appropriateness of explanations.

The process model is straightforward. First, we answer the \textit{`why?'} and \textit{`who?'} for the domain to identify the list of potential explainees and the goals of explanation for each of these.  The layers are used to answer the \textit{`how?'}, where the answer to these involves identifying which layer is appropriate$^*$.

\marginpar{
$^*$To ensure that the outcome of these questions remains impartial, we recommend using the idea of the \textit{Original Position} (OP), proposed by political philosopher John Rawls \cite{Rawls1971}. The ``most appropriate moral conception of justice'' \cite{Freeman2019OP} is obtained when the parties take up the ``\textit{veil of ignorance}'', completely depriving themselves of all knowledge of their own personal circumstances and attributes; in short, putting themselves in the shoes of others.
}

\subsection{High stakes example 1: AI-based credit scoring}
The first example is for a typical algorithmic decision-making case study \cite{Chen2018}. Consider a case in which the `who?' question is an applicant whose loan application has been rejected, and the `why?' is for them to learn how they can get approval in future. The different layers are:

\begin{description}
\item[Baseline] No explanation: `Your loan has been denied'.
\item[Layer 1] Feature-based: `The decision was made based on these variables: income, with these weights: \ldots'.
\item[Layer 2] Contrastive: `The loan may be approved if the applicant were to have an income of more than \ldots'.
\item[Layer 3] Directive: `The loan may be approved if the applicant were to have an income of more than \ldots. The applicant could get a second job or ask for a promotion to increase the income.'
\end{description}

Credit risk assessment is a high stakes domain. We know from earlier works~\cite{Binns2018,singh2021directive} some explanations are deemed inappropriate. While \emph{income} is a legitimate feature for the AI to use, its use can be offensive depending on \emph{how} it is used in an explanation. As such, we may decide to provide explanations up to Layer 2 (contrastive). However, explaining is a social or interactive process~\cite{miller2019explanation}, and we could increase the \emph{level of verbosity} (go to Layer 3) if the recipient requests further details.

\subsection{Low stakes example: IT Admin assisting an employee with a disabled user account}
In the following example, the domain and feature are both in the lower quadrants; we could offer Layer 3 explanation.
\begin{description}
\item[Layer 1] `Account disabled due to login attempts.'
\item[Layer 2] `Account disabled due to more than 10 login attempts.'
\item[Layer 3] `Account disabled due to more than 10 login attempts. To resolve, either reset the login attempts counter or wait for 48 hours for an automatic reset.'
\end{description}

\section{Final Remarks}
While \emph{who} we explain to is critical~\cite{ehsan2020human}, \emph{how} we explain is equally important. Studies~\cite{Binns2018,singh2021directive} reveal that current XAI tools present inappropriate explanations to people when using certain features. Building on the \emph{layers of explanation} (LEx) framework that enables practitioners to judge an explanation's appropriateness, we need to further explore the following aspects:

\begin{enumerate}
    \item
       What factors make an attribute sensitive for an individual, and how should we operationalise these factors in an explanation generation pipeline to align with one's explanatory needs?
    \item 
        How do these sensitivity factors change with application domain?
    \item 
        What design guidelines should we incorporate during model development and deployment pipeline to empower users with the agency to decide the \textit{what, when,} and \textit{how} of an explanation?
        
\end{enumerate}

By tackling these questions, we can refine the bounds of explainability in a human-centered manner, one that accommodate societal expectations and cater to the human factors that govern people's reaction to AI-mediated explanations.  

\bibliographystyle{SIGCHI-Reference-Format}
\bibliography{references}

\end{document}